\documentclass[letterpaper, 10 pt, conference]{ieeeconf}  % Comment this line out if you need a4paper
\usepackage{amsmath,amsfonts}
\usepackage{algorithm}
\usepackage{algorithmicx}
\usepackage{algpseudocode}
\usepackage{array}
\usepackage{textcomp}
\usepackage{stfloats}
\usepackage{url}
\usepackage{verbatim}
\usepackage{graphicx}
\usepackage{tabularx}
\usepackage{siunitx}
\usepackage{multirow}
\usepackage{array}
\usepackage{amssymb}  
\usepackage{booktabs}
\usepackage{upgreek}
\usepackage{graphicx}
\usepackage{subfig}
\usepackage{amsmath}
\usepackage{amsthm} % definte the proof environment: block letters + italic letters
\usepackage{amssymb}
\usepackage[font=footnotesize]{caption} % set the captain font size to 8 (i.e. footnotesize)
\usepackage{subfig} % uses subfloats within a single float MUST after the package {caption}!!
\usepackage[noadjust]{cite} % alphabetize grouped citations automatically [3, 1, 2] to [1, 2, 3]
\usepackage{color}
\usepackage{algorithm} % options: boxed [section]
\usepackage{algpseudocode} % for algorithm
\usepackage{enumerate}
\usepackage[hidelinks,colorlinks=false]{hyperref}
\usepackage[left=0.75in, right=0.75in, top=0.75in, bottom=0.75in]{geometry}
\usepackage{authblk} % for authors' affiliation \affil{xxx}
\usepackage{arydshln} % for dashline in table or matrix
\usepackage{subfloat}
\usepackage{mdframed}
\usepackage{bm}
\usepackage{tikz}
\usetikzlibrary{calc} % for calculation functions in Tikz let, in commands in Tikz
\usetikzlibrary{shapes} % for block diagram
\usetikzlibrary{chains}
\usetikzlibrary{fit}
\usetikzlibrary{arrows}
\usetikzlibrary{decorations.text} % text along path
\usetikzlibrary{decorations.markings}
\usetikzlibrary{decorations.pathmorphing} % for zigzag arrow
\usetikzlibrary{shadows}
\usetikzlibrary{patterns}
\usetikzlibrary{matrix}
\usepackage{pgfplots}
\usepackage[europeanresistors]{circuitikz}
\usepackage[outline]{contour} % white surround text
\contourlength{1.5pt}
\IEEEoverridecommandlockouts

\title{\LARGE \bf
Multi-Task Reinforcement Learning of Drone Aerobatics by Exploiting Geometric Symmetries
}
\author{Zhanyu Guo$^{1,2}$, Zikang Yin$^{1,2}$, Guobin Zhu$^{3,2}$, Shiliang Guo$^{2}$, and Shiyu Zhao$^{2*}$% <-加一个*号方便对应通讯作者
\thanks{*Corresponding author}%
\thanks{$^{1}$College of Computer Science and Technology, Zhejiang University, Hangzhou, China.}%
\thanks{$^{2}$WINDY Lab, School of Engineering, Westlake University, Hangzhou, China. 
        {\{guozhanyu, yinzikang, guoshiliang, zhaoshiyu\}@westlake.edu.cn}}%
\thanks{$^{3}$School of Automation Science and Electrical Engineering, Beihang University, Beijing, China.
        {zhugb@buaa.edu.cn}}%
\thanks{This work was supported by the Brain Science and Brain-like Intelligence Technology — National Science and Technology Major Project (Grant No. 2022ZD0208800) and National Natural Science Foundation of China (Grant No. 62473320)}%
}

\begin{document}

\bstctlcite{IEEEexample:BSTcontrol}

\maketitle
\thispagestyle{empty}
\pagestyle{empty}
\begin{abstract}
Flight control for autonomous micro aerial vehicles (MAVs) is evolving from steady flight near equilibrium points toward more aggressive aerobatic maneuvers, such as flips, rolls, and Power Loop. Although reinforcement learning (RL) has shown great potential in these tasks, conventional RL methods often suffer from low data efficiency and limited generalization. This challenge becomes more pronounced in multi-task scenarios where a single policy is required to master multiple maneuvers. In this paper, we propose a novel end-to-end multi-task reinforcement learning framework, called GEAR (Geometric Equivariant Aerobatics Reinforcement), which fully exploits the inherent SO(2) rotational symmetry in MAV dynamics and explicitly incorporates this property into the policy network architecture. By integrating an equivariant actor network, FiLM-based task modulation, and a multi-head critic, GEAR achieves both efficiency and flexibility in learning diverse aerobatic maneuvers, enabling a data-efficient, robust, and unified framework for aerobatic control. GEAR attains a 98.85\% success rate across various aerobatic tasks, significantly outperforming baseline methods. In real-world experiments, GEAR demonstrates stable execution of multiple maneuvers and the capability to combine basic motion primitives to complete complex aerobatics.
\end{abstract}

\section{Introduction}

Recent years have witnessed a paradigm shift in autonomous aerial robotics, moving beyond simple navigation towards highly dynamic and aggressive aerobatic maneuvers such as Multi-Flip and Power Loop\cite{kaufmann2020RSS,wang_unlocking_2025,han2025reactive}.
This leap in agility unlocks critical capabilities for rapid-response and high-precision applications, including autonomous free-style flight~\cite{wang_unlocking_2025} and drone racing~\cite{kaufmann_champion-level_2023}. 
Learning-based methods have been at the forefront of this push, achieving or even surpassing human-level performance in complex flight tasks~\cite{song2023reaching}.

Current research in autonomous drone acrobatics follows mainly two paradigms. 
The first paradigm is the conventional planning-then-tracking approach~\cite{wang_unlocking_2025}. 
This method uses differential flatness and spatial-temporal joint optimization to generate feasible trajectories. 
Controllers like Model Predictive Control (MPC) or geometric controllers then track these trajectories. 
This approach allows drones to perform complex maneuvers and ensures safety in cluttered environments. 
However, it faces computational challenges that prevent real-time replanning. 
It also struggles with singularities at extreme attitudes, which makes it less suitable for aggressive aerobatic flight.

The second paradigm is the data-driven approach, which directly learns a mapping from state and intention to low-level commands~\cite{han2025reactive, wang_aerobatic_2022, kaufmann2020RSS, yin_taco_2025}. 
Early works adopted imitation learning, requiring extensive expert demonstrations to achieve human-level acrobatics, but struggled with generalization and scalability. 
Recent advances employ RL to autonomously discover agile flight policies. 
Some methods focus on mastering specific maneuvers such as Tic-Toc or spin by rewarding the physical properties of acrobatics~\cite{wang_aerobatic_2022}, while others introduce waypoint-based intention encoding to enable robust tracking of sequential aerobatic waypoints in large-scale and dynamic scenarios~\cite{han2025reactive}. 
However, these RL systems typically rely on carefully engineered waypoint sequences~\cite{han2025reactive}, restricting spontaneity.

To transcend these limitations, the multi-task reinforcement learning (MTRL) is the logical next step. MTRL has proven to be effective in learning policies in related tasks using their shared structure such as  Feature-wise Linear Modulation (FiLM) layer and multihead network  ~\cite{xing2024multi,bauersfeld_user-conditioned_2023}. In MAV control, MTRL has been used to develop unified policies for diverse behaviors such as stabilization, trajectory tracking, and aggressive racing, conditioned on the command input~\cite{xing2024multi,zhu_equact_2025}. 
But this approach dramatically exacerbates the data-efficiency problem, making it prohibitively expensive with standard methods~\cite{xing2024multi,zhu_equact_2025}. 
The central challenge, therefore, is to create an MTRL framework that is data-efficient enough to master multiple high-speed maneuvers simultaneously.

This data-efficiency challenge stems from the fact that standard RL methods must learn about the underlying system dynamics from scratch, including fundamental physical symmetries. 
To solve this problem, equivariant learning, grounded in geometric learning principles, offers a promising direction for enhancing data efficiency by explicitly encoding domain-specific symmetries—such as translations, rotations, and permutations into neural network architectures~\cite{cesa2022a}. 
This approach improves sample efficiency in vision tasks~\cite{cesa2022a}, vision-based robotic manipulation~\cite{huang2023leveraging}. 
While this technique has been applied to simpler MAV tracking tasks, its potential to enable complex, multi-task acrobatic control has not been explored~\cite{yu_equivariant_2025}.

However, recent research has shown that strictly enforcing equivariance may limit policy expressiveness and robustness in diverse or asymmetric environments~\cite{pertigkiozoglou2024improving,NEURIPS2023_c95c0496,wang_practical_2025}. To address this limitation, some approaches relax or partially remove symmetry constraints during training, which stabilizes optimization~\cite{pertigkiozoglou2024improving}, while others allow controlled symmetry breaking at test time so that policies can adapt flexibly to task-specific goals~\cite{NEURIPS2023_c95c0496}. Recent advances combine invariant action representations with non-equivariant policy heads, or leveraging pretrained encoders with approximate symmetrization in diffusion policies, show that selectively breaking symmetry can achieve performance comparable to or even surpassing fully equivariant models~\cite{wang_practical_2025,zhu_equact_2025}. Collectively, these studies suggest that symmetry is best treated as a soft inductive bias: relaxing or breaking it where necessary yields policies that are more expressive and robust, while preserving the efficiency gains of symmetry where appropriate.

\noindent\textbf{Contribution:}  
The contributions and novelties of this work are as follows. To enable efficient learning of diverse aerobatic maneuvers without relying on manually designed waypoint sequences, we propose a unified multi-task reinforcement learning framework called \textit{GEAR} (Geometric Equivariant Aerobatics Reinforcement). GEAR explicitly embeds the inherent SO(2) rotational symmetry of MAV dynamics into the policy architecture. The actor network employs an SO(2)-equivariant backbone to exploit geometric priors, while Feature-wise Linear Modulation (FiLM) provides lightweight task conditioning, enabling the same backbone to flexibly represent different maneuvers. In parallel, a multi-head critic delivers task-specific value estimation, ensuring that learning signals remain well separated across heterogeneous objectives. This design allows GEAR to efficiently capture shared structures across tasks, while adapting to task-specific variations, thereby addressing the long-standing challenge of data inefficiency in multi-task aerobatic learning.  

By combining symmetry-aware representations with adaptive modulation, GEAR enables a single end-to-end policy to master multiple high-speed aerobatic maneuvers. To the best of our knowledge, this is the first end-to-end multi-task RL framework that explicitly encodes SO(2) geometric symmetry into the policy architecture. To validate GEAR, we first conduct high-fidelity simulation experiments, where GEAR achieves a final training return 9.53\% higher than baseline methods and reaches a 98.85\% success rate across various aerobatic tasks, significantly outperforming ablation variants. We then deploy the learned policy on physical MAV platforms. Real-world experiments demonstrate that a single policy can execute diverse high-speed maneuvers with adjustable parameters (e.g., flip speed, number of rolls, rotation velocity). By composing the four basic primitives, we successfully realize complex aerobatics such as Power Loop, Barrel Roll, and Multi-Flip, showcasing both the robustness and practicality of the proposed approach.

\section{Related Work}
\subsection{MAVs' Acrobatic Maneuvers}
The pursuit of aerobatic flight in aerial robotics has inspired a wide spectrum of research efforts~\cite{kaufmann2020RSS,wang_unlocking_2025}. 
Prior work has explored diverse challenges, ranging from discrete, high-dynamic maneuvers such as flips~\cite{han2025reactive}, Power Loop~\cite{wang_unlocking_2025}, or Tic-Tocs~\cite{wang_aerobatic_2022}, to trajectory-based tasks such as racing through complex tracks and performing freestyle sequences in cluttered environments~\cite{wang_unlocking_2025,kaufmann_champion-level_2023}. 
Control strategies have evolved from classical model-based approaches, which provide stability guarantees when accurate models are available~\cite{sun_comparative_2022}, to learning-based methods. 
In particular, RL has enabled end-to-end policy learning, achieving or even surpassing human-level performance in complex aerobatic tasks~\cite{kaufmann_champion-level_2023,wang_unlocking_2025,yin_taco_2025,han2025reactive,kaufmann2020RSS}. 

\subsection{Multi-Task Reinforcement Learning for Robotics}
To enhance the versatility of a single agent, MTRL has emerged as a powerful paradigm~\cite{zhao2025RLBook,xing2024multi,bauersfeld_user-conditioned_2023,UNMANNED}. 
MTRL aims to jointly learn multiple related tasks, allowing knowledge transfer between tasks to improve both generalization and data efficiency. 
A central challenge lies in managing potentially conflicting objectives, which can sometimes degrade performance compared to training separate specialized agents~\cite{bauersfeld_user-conditioned_2023}. Recent work has applied MTRL to MAV control, demonstrating that a single policy can be trained for distinct flight behaviors such as stabilization, velocity tracking, and racing by leveraging specialized architectures such as multiple critics and shared encoders~\cite{xing2024multi}. 
Nevertheless, extending MTRL to complex, high-speed aerobatics dramatically increases data requirements, often making training prohibitively expensive with standard methods~\cite{xing2024multi}. 
Our work addresses this efficiency challenge directly.

\subsection{Equivariant Learning}
Equivariant learning, grounded in geometric deep learning principles, encodes domain-specific symmetries into neural network architectures~\cite{cesa2022a}. This approach has been successfully applied to robotic manipulation~\cite{huang2023leveraging,simeonov2022neural,wang2022so}, legged locomotion~\cite{su_leveraging_2024,ordonez-apraez_morphological_2025}. For MAVs, symmetry-aware policies have been explored, but existing studies are largely limited to relatively simple tasks such as hovering or trajectory tracking~\cite{smith_so2-equivariant_2024,yu_equivariant_2025,yu_equivariant_2023,10054413}. The potential of equivariant learning to enable a data-efficient, unified MTRL policy for complex aerobatics remains largely unexplored. Recent work has shown that strictly enforcing equivariance can reduce policy expressiveness and robustness in diverse or asymmetric environments~\cite{pertigkiozoglou2024improving,NEURIPS2023_c95c0496,wang_practical_2025}. To address this, these approaches relax or partially remove symmetry constraints during training or testing, highlighting that symmetry is best viewed as a soft inductive bias—retaining its efficiency benefits where applicable while relaxing or breaking it when necessary to enhance generalization and robustness.

\section{Problem Formulation}

\subsection{MAV Dynamics}

Throughout this paper, all vectors are treated as column vectors, and transpose symbols $^\top$ are omitted when the vector shape is clear from context. 
The MAV's state is characterized by position $\mathbf{p} \in \mathbb{R}^3$, linear velocity $\mathbf{v} \in \mathbb{R}^3$, attitude $\mathbf{R} \in \mathrm{SO}(3)$, and angular velocity $\boldsymbol{\omega} \in \mathbb{R}^3$.  
The governing equations are
\begin{equation}
\begin{aligned}
\dot{\mathbf{p}} &= \mathbf{v}, 
& \quad 
m\,\dot{\mathbf{v}} &= m\,\mathbf{g} + \mathbf{R}\,\mathbf{e}_3 f_\Sigma - \mathbf{K}_{\text{drag}}\,\mathbf{v}, \\
\dot{\mathbf{R}} &= \mathbf{R}[\boldsymbol{\omega}]_\times, 
& \quad 
\dot{\boldsymbol{\omega}} &= \mathbf{J}^{-1} \bigl(\boldsymbol{\tau} - [\boldsymbol{\omega}]_\times \mathbf{J} \boldsymbol{\omega}\bigr).
\end{aligned}
\end{equation}

where $\mathbf{e}_3 = [0, 0, 1]^\top$, $\mathbf{g} = [0, 0, -9.81]^\top$, $m>0$ and $\mathbf{J} \in \mathbb{R}^{3\times3}$ are the mass and inertia, and $\mathbf{K}_\text{drag}\in \mathbb{R}^{3\times3}$ is a diagonal drag coefficient matrix. The total thrust $f_\Sigma$ and torque $\boldsymbol{\tau}$ are computed as
\begin{align}
\begin{split}
f_\Sigma &= \sum_{i=1}^{4} f_i,
\end{split}
\;
\begin{split}
\boldsymbol{\tau} &= \sum_{i=1}^{4} (\mathbf{r}_i \times \mathbf{e}_3 f_i + \tau_i),
\end{split}
\end{align}
where $f_i \in \mathbb{R}^3$ and $\tau_i\in \mathbb{R}^3$ are the thrust and torque generated by motor $i$, $\mathbf{r}_i$ is the arm of force.

\subsection{State Design}
In the MTRL setting, we consider $N$ tasks, each defined by an MDP $\mathcal{M}_i = (\mathcal{S}_i, \mathcal{A}_i, \mathcal{P}_i, \mathcal{R}_i, \gamma_i)$, with shared dynamics and task-specific rewards. 
The optimization objective is
$
    J(\pi) = \frac{1}{N} \sum_{i=1}^{N} \mathbb{E}_{\pi} \left[ \sum_{t=0}^{\infty} \gamma_i^t r_i(s_t, a_t) \right].
$
To ensure generality across basic hovering and diverse maneuvers, the state is designed to consist of three components:  
(i) the relative state $\mathbf{s}_{\mathrm{rel}}$, which characterizes task-invariant relationships,  
(ii) the previous action $\mathbf{a} = \left[f_\Sigma,\, \boldsymbol{\omega}\right] \in \mathbb{R}^4$, included to mitigate delays caused by discrete sampling, and  
(iii) the command input $cmd \in \mathbb{R}^5$, which encodes the maneuver type (one-hot) together with a scalar parameter specifying its attribute.  

The relative state $\mathbf{s}_{\mathrm{rel}}$ in the body frame is defined as  
\begin{align}
\mathbf{s}_{\mathrm{rel}} &= \left[ \mathbf{p}_{\mathrm{rel}},\, \mathbf{v}_{\mathrm{rel}},\, 
\boldsymbol{\omega}_{\mathrm{rel}},\, \mathbf{R}_{\mathrm{rel}} \right] \in \mathbb{R}^{18}, 
\end{align}
\begin{equation}
\begin{aligned}
\mathbf{p}_{\mathrm{rel}} &= \mathbf{R}^\top (\mathbf{p}_{\mathrm{des}} - \mathbf{p}), 
& \quad \mathbf{v}_{\mathrm{rel}} &= \mathbf{R}^\top (\mathbf{v}_{\mathrm{des}} - \mathbf{v}), \\
\boldsymbol{\omega}_{\mathrm{rel}} &= \mathbf{R}^\top (\boldsymbol{\omega}_{\mathrm{des}} - \boldsymbol{\omega}), 
& \quad \mathbf{R}_{\mathrm{rel}} &= \mathbf{R}^\top \mathbf{R}_z(\psi_{\mathrm{des}}),
\end{aligned}
\label{eq:rel-state-sd}
\end{equation}
where the desired position $\mathbf{p}_{\mathrm{des}}$, yaw $\psi_{\mathrm{des}}$, linear velocity $\mathbf{v}_{\mathrm{des}}$, and angular velocity $\boldsymbol{\omega}_{\mathrm{des}}$ define the task-specific invariants.  
The yaw $\psi_{\mathrm{des}}$ does not prescribe a particular heading but instead fixes the orientation of the maneuver plane in the world frame.  
This ensures that the maneuvers are yaw-equivariant: their definitions remain valid under arbitrary global yaw rotations.

\noindent\textbf{Hover:}  
In Hover task, the MAV must reach and maintain a specified position $\mathbf{p}_{\mathrm{des}}$ with yaw $\psi_{\mathrm{des}}$, while desired velocities are $\mathbf{v}_{\mathrm{des}} = [0,0,0]$ and $\boldsymbol{\omega}_{\mathrm{des}} = [0,0,0]$.  
Task success corresponds to $\mathbf{p}_{\mathrm{rel}} = [0,0,0]$, $\mathbf{R}_{\mathrm{rel}} = \mathbf{I}_3$, $\mathbf{v}_{\mathrm{rel}} = [0,0,0]$, and $\boldsymbol{\omega}_{\mathrm{rel}} = [0,0,0]$, regardless of the global yaw.  
This configuration serves as both a safe initialization and a termination condition for aerobatic maneuvers.  

\noindent\textbf{Rotate:}  
In the Rotate maneuver, the MAV orbits a fixed target in the horizontal plane with radius $r$ and tangential velocity $v$.  
In the body frame, the target remains at a constant relative position, yielding invariants $\mathbf{p}_{\mathrm{des}} = [r,0,0]$ and $\mathbf{v}_{\mathrm{des}} = [0,v,0]$.  

\noindent\textbf{Flip:}  
In the Flip maneuver, the MAV performs a vertical rotation about its body $y$-axis, moving along a circular trajectory of radius $r$ in the $x$–$z$ plane aligned with yaw $\psi_{\mathrm{des}}$.  
The motion is parameterized by a commanded angular velocity $\omega$, leading to desired invariants $\mathbf{p}_{\mathrm{des}} = [0,0,r]$, $\mathbf{v}_{\mathrm{des}} = [\omega r,0,0]$, and $\boldsymbol{\omega}_{\mathrm{des}} = [0,\omega,0]$.  
Here, $\psi_{\mathrm{des}}$ anchors the orientation of the flip plane in the world frame, decoupling the maneuver from the global yaw.  

\noindent\textbf{Roll:}  
In the Roll maneuver, the MAV rotates about its body $x$-axis while remaining centered at $\mathbf{p}_{\mathrm{des}}$ (i.e., $r=0$).  
Unlike the Flip, the Roll is specified by a commanded number of turns $N$ instead of a fixed angular velocity.  
The cumulative roll angle $\phi$ is tracked modulo $2\pi$, and the task terminates when $\phi = 2\pi N$.  
The Roll occurs in the $y$–$z$ plane aligned with yaw $\psi_{\mathrm{des}}$, ensuring that the definition of the maneuver remains invariant to changes in global yaw.

\subsection{Reward Function Design}
A unified reward formulation is adopted across all maneuver tasks, such that the policy learns from a consistent set of performance indicators while preserving task-specific distinctions.  
The overall reward is defined as
\begin{equation}
r = r_{\text{pos}} \cdot r_{\text{lin}} \cdot r_{\text{ang}} \cdot r_{\text{cmd}} \cdot r_{\text{task}},
\end{equation}
where $r_{\text{pos}}$, $r_{\text{lin}}$, and $r_{\text{ang}}$ are basic tracking terms based on relative quantities, $r_{\text{cmd}}$ encourages adherence to high-level commands, and $r_{\text{task}}$ introduces maneuver-specific shaping. All reward components are shaped by the kernel $H(x; k) = \frac{1}{1 + kx}$, where $k>0$ is a sensitivity factor. Small values of $k$ yield smooth shaping, while larger values impose sharper penalties for deviations.  

\paragraph{Basic tracking terms}
We impose simultaneous constraints on position, linear velocity, and angular velocity:
\begin{align}
r_{\text{pos}} &= \sum_{k \in \{1,10\}} H(\|\mathbf{p}_{\mathrm{rel}}\|^2; k), \\
r_{\text{lin}} &= \sum_{k \in \{1,10,100\}} H(\|\mathbf{v}_{\mathrm{rel}}\|^2; k), \\
r_{\text{ang}} &= \sum_{k \in \{0.1,1,10\}} H(\|\boldsymbol{\omega}_{\mathrm{rel}}\|^2; k).
\end{align}

These terms ensure that the MAV cannot exploit single dimensions of the state and must maintain accurate tracking in all channels.

\paragraph{Command adherence}
To guarantee that the learned policy respects high-level maneuver commands, we introduce
\begin{equation}
r_{\text{cmd}} = \sum_{k \in \{1,10\}} H\bigl((a_{\mathrm{ach}} - a_{\mathrm{cmd}})^2; k\bigr),
\end{equation}
where $a_{\mathrm{cmd}}$ denotes the commanded attribute (e.g., angular velocity, number of turns), and $a_{\mathrm{ach}}$ is its achieved counterpart measured during execution.

\paragraph{Task-specific shaping}
Each maneuver imposes an additional geometric or orientation requirement beyond the basic tracking terms.  
\noindent\emph{Hover:} Orientation alignment is encouraged via quaternion similarity, expressed as  
$r_{\text{task}} = \sum_{k \in \{1,10\}} H\!\left(2 \arccos(|q \cdot q_{\mathrm{des}}|); k\right)$,  
where $q$ and $q_{\mathrm{des}}$ are the current and desired unit quaternions.  \noindent\emph{Rotate:} Since the target remains at a constant location in the body frame, success corresponds to maintaining a vanishing lateral offset:  
$r_{\text{task}} = \sum_{k \in \{0.1,1\}} H\!\left(p_{\mathrm{rel},y}^2; k\right)$,  
where $p_{\mathrm{rel},y}$ denotes the $y$-component of $\mathbf{p}_{\mathrm{rel}}$.  \noindent\emph{Roll:} Alignment of the body $x$-axis with the desired roll axis is promoted using the body-to-world rotation matrix $\mathbf{R}_{bw}$, with  
$r_{\text{task}} = \sum_{k \in \{1,10\}} H\!\left((1 - \mathbf{R}_{bw}[0,0])^2; k\right)$.  \noindent\emph{Flip:} a constant reward is assigned,  
$r_{\text{task}} = 2$.

\subsection{SO(2)-Symmetry and Equivariant Structure}

A fundamental property of MAV dynamics is its invariance under rotations about the gravitational axis. 
This $SO(2)$ symmetry implies that the drone's physical behavior and control performance are independent of the yaw orientation in the horizontal plane in world frame~\cite{yu_equivariant_2023}. 

\paragraph{Group action and representations}
Formally, let the symmetry group $\mathbb{G} = SO(2)$ act on the state space
\begin{equation}
\mathcal{S} = (\mathbf{p}, \mathbf{v}, \mathbf{R}, \boldsymbol{\omega}) \in \mathbb{R}^3 \times \mathbb{R}^3 \times SO(3) \times \mathbb{R}^3.
\end{equation}
For $\theta \in [0, 2\pi)$, the group action $g_\theta \in \mathbb{G}$ is
\begin{equation}
    g_\theta \circ s = 
    \big( \mathbf{R}_z(\theta)\mathbf{p},\; \mathbf{R}_z(\theta)\mathbf{v},\; \mathbf{R}_z(\theta)\mathbf{R},\; \boldsymbol{\omega} \big),
    \label{eq:state_representation}
\end{equation}
where $\mathbf{R}_z(\theta)$ is the rotation about the world $z$-axis. 
Here $\mathbf{p}$ and $\mathbf{v}$ rotate in the world frame, $\mathbf{R}$ is left-multiplied, while the body-rate $\boldsymbol{\omega}$ (defined in the body frame) remains unchanged. 
The action space $\mathcal{A} = [f_{\Sigma}, \boldsymbol{\omega}]$ is body-frame defined and represented by the trivial action $\rho_{\mathcal{A}}(\theta)=\mathrm{diag}(1, \mathbf{I}_3)$, where $
\rho_{\mathcal{A}} : SO(2) \rightarrow GL(\mathcal{A}).
$

\paragraph{Equivariance of dynamics}
The MAV dynamics function $F: \mathcal{S} \times \mathcal{A} \to T\mathcal{S}$ is \emph{equivariant} under this action if (i) gravity is aligned with the world $z$-axis, and (ii) the drag matrix is isotropic in the $x$-$y$ plane ($K_{xx}=K_{yy}$). 
Under these conditions,
\begin{equation}
    F(g_\theta \circ s, a) = g_\theta \circ F(s, a), \qquad\forall g \in \mathbb{G}, \forall \theta,
    \label{eq:equivariance_formal}
\end{equation}
so that for any trajectory $(s(t), a(t))$, the rotated trajectory $(g_\theta \circ s(t), a(t))$ also satisfies the dynamics.

\paragraph{Equivariance in body-frame representations}
Although $SO(2)$ symmetry is defined in the world frame, our policy uses the body-frame relative state in~\eqref{eq:rel-state-sd}. 
Under a global yaw rotation $\mathbf{R}_z(\theta)$, the MAV state becomes $(\mathbf{R}_z\mathbf{p},\mathbf{R}_z\mathbf{v},\mathbf{R}_z\mathbf{R},\boldsymbol{\omega})$, and the desired targets become $(\mathbf{R}_z\mathbf{p}_{\mathrm{des}},\mathbf{R}_z\mathbf{v}_{\mathrm{des}},\mathbf{R}_z\boldsymbol{\omega}_{\mathrm{des}},\psi_{\mathrm{des}}+\theta)$. Substituting these into~\eqref{eq:rel-state-sd} shows invariance:
$
\mathbf{p}'_{\mathrm{rel}} = (\mathbf{R}_z\mathbf{R})^\top(\mathbf{R}_z\mathbf{p}_{\mathrm{des}}-\mathbf{R}_z\mathbf{p}) = \mathbf{R}^\top(\mathbf{p}_{\mathrm{des}}-\mathbf{p}) = \mathbf{p}_{\mathrm{rel}},
$
$
\mathbf{v}'_{\mathrm{rel}} = (\mathbf{R}_z\mathbf{R})^\top(\mathbf{R}_z\mathbf{v}_{\mathrm{des}}-\mathbf{R}_z\mathbf{v}) = \mathbf{v}_{\mathrm{rel}}, $
$\boldsymbol{\omega}'_{\mathrm{rel}} = (\mathbf{R}_z\mathbf{R})^\top(\mathbf{R}_z\boldsymbol{\omega}_{\mathrm{des}}-\boldsymbol{\omega}) = \boldsymbol{\omega}_{\mathrm{rel}},
$
$
\mathbf{R}'_{\mathrm{rel}} = (\mathbf{R}_z\mathbf{R})^\top \mathbf{R}_z(\psi_{\mathrm{des}}+\theta)
= \mathbf{R}^\top\mathbf{R}_z(\psi_{\mathrm{des}}) = \mathbf{R}_{\mathrm{rel}}.
$
Thus, all relative components remain unchanged, and the world-frame $SO(2)$ symmetry is faithfully preserved in the body-frame representation used as policy inputs.

\paragraph{Task symmetry analysis}
Not all tasks interact with this symmetry in the same way.  
\noindent\emph{Hover:} Enforces an absolute yaw $\mathbf{R}_{\mathrm{des}}$ in the world frame, thereby breaking $SO(2)$ invariance.  
\noindent\emph{Roll:} Specifies rotation about the body $x$-axis, independent of global yaw. 
\noindent\emph{Flip:} Commands angular velocity about a body-fixed axis; rotating the system about $z$ yields an equivalent trajectory.  
\noindent\emph{Rotate:} Defines tangential motion around a target; the relative target position $\mathbf{p}_{\mathrm{rel}}$ is yaw-invariant.  
In summary, Hover introduces deliberate symmetry breaking, whereas Roll, Flip, and Rotate preserve the underlying $SO(2)$ symmetry both in world-frame dynamics and in body-frame state representations.

\subsection{Equivariant Reinforcement Learning}

If the transition $T(\cdot)$, initial state $p_0(\cdot)$, and reward $r_i(\cdot)$ are $G$-invariant, i.e.,
\begin{align} 
T(g \circ s' \mid g \circ s, a) &= T(s' \mid s, a), \\
p_0(g \circ s) &= p_0(s), \\
r_i(g \circ s, a) &= r_i(s, a),
\end{align}
then the optimal value function and policy also satisfy
\begin{equation}
V^*(g \circ s) = V^*(s), \qquad \pi^*(g \circ s) = \pi^*(s).
\end{equation}
Because actions are expressed in the body frame, the group action reduces to the identity, so the optimal action is in fact invariant under $SO(2)$ rotations~\cite{yu_equivariant_2023,yu_equivariant_2025}. 

In multi-task learning, however, a strictly equivariant network may over-constrain learning, as different maneuvers impose different symmetry requirements. 
Furthermore, reward design compounds this challenge: beyond a command term enforcing consistency with high-level inputs, each maneuver introduces task-specific rewards emphasizing distinct geometric or dynamic properties. 
While such heterogeneous rewards are essential for success, they bias the policy toward different feature subsets, limiting the generalization of a strictly equivariant representation. 
This tension between symmetry, reward shaping, and task diversity motivates a more flexible framework that retains the efficiency of equivariance while adapting to maneuver-specific asymmetries.

\section{Proposed Method}
\begin{figure*}[t]
    \centering
    \includegraphics[width=1\textwidth]{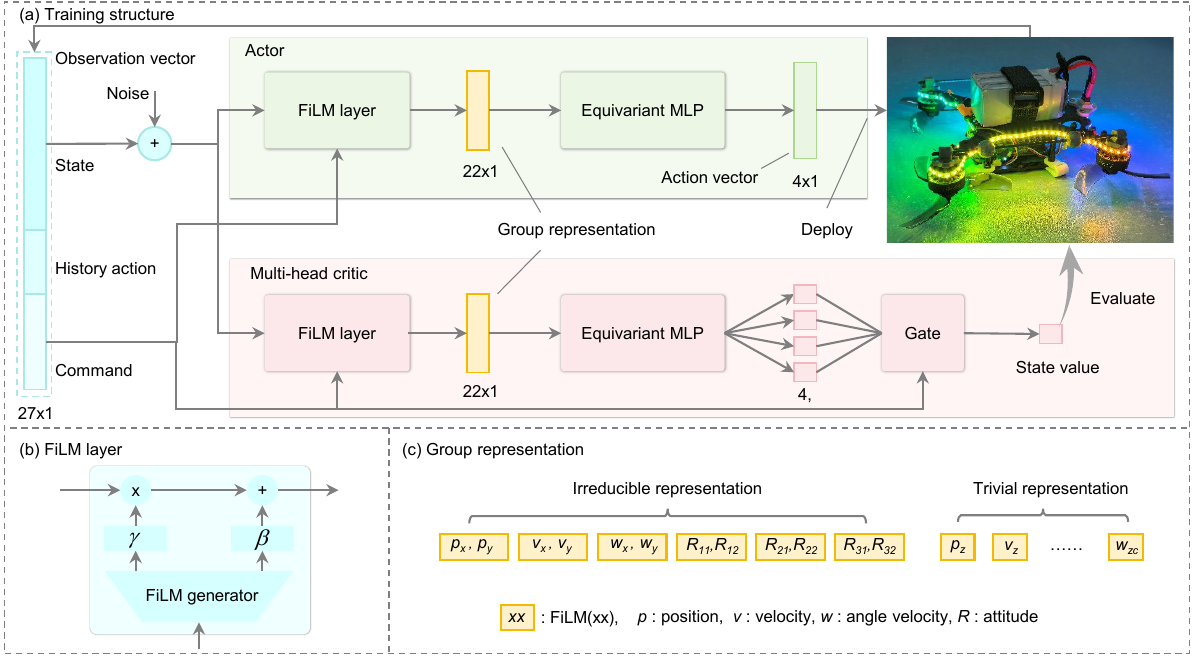}
\caption{Overview of the proposed GEAR framework. 
(a) Training structure: The policy consists of an actor and a multi-head critic, both combining FiLM layers with an Equivariant MLP (EMLP). 
The actor receives state(with added noise during training), command, and history action inputs and outputs low-level control actions. 
The critic provides task-specific value estimates, which are used only in training and discarded at deployment. 
(b) FiLM layer: commands generate scaling and shifting factors ($\gamma$, $\beta$) to modulate intermediate features, enabling task conditioning and controlled symmetry breaking. 
(c) Group representation: state features are partitioned into irreducible and trivial components.}

\label{fig:flowchart}
\end{figure*}

The proposed GEAR framework integrates geometric priors, command conditioning, and modular value estimation into a unified architecture for multi-task aerobatics. 
An overview is shown in Fig.~\ref{fig:flowchart}. 
The framework consists of an SO(2)-equivariant actor network, FiLM-based task conditioning, and a multi-head critic. 
During training, the actor receives noisy observations along with command inputs and outputs low-level control actions, while the critic provides task-specific value estimates that guide optimization. 
The FiLM layers allow task-dependent modulation of the shared equivariant backbone, enabling controlled symmetry breaking where necessary. 
At deployment, only the actor is retained, and the learned primitives (Hover, Flip, Roll, Rotate) can be composed into complex aerobatic maneuvers. 
\subsection{Group Representation}
The input state is organized according to its transformation properties under $SO(2)$. 
Specifically, (i) \emph{equivariant pairs}, such as $(p_{\mathrm{rel},x},p_{\mathrm{rel},y})$, $(v_{\mathrm{rel},x},v_{\mathrm{rel},y})$, and $(\omega_{\mathrm{rel},x},\omega_{\mathrm{rel},y})$, transform under the 2D irreducible representation $\rho_{\mathrm{vec}}: SO(2) \to \mathbb{R}^{2\times 2}$, defined as the planar rotation matrix
$\rho_{\mathrm{vec}}(\theta)=\mathbf{R}(\theta)$; 
(ii) \emph{invariants}, including $p_{\mathrm{rel},z}, v_{\mathrm{rel},z}, \omega_{\mathrm{rel},z}, f_\Sigma$, and scalar command inputs, follow the trivial representation $\rho_{\mathrm{sca}}: SO(2) \to \mathbb{R}$, defined as $\rho_{\mathrm{sca}}(\theta)=1$.
The overall group action on the input is therefore $\mathrm{diag}(\rho_{\mathrm{vec}}^{\oplus }, \rho_{\mathrm{sca}}^{\oplus })$, while the action space, expressed in the body frame, is invariant. 
This setup ensures that rotation-sensitive features transform consistently, while invariant channels remain fixed. 
FiLM layers applied before grouping allow controlled symmetry breaking when task asymmetries are present.

\subsection{SO(2)-Equivariant Actor Network}

The actor network adopts an EMLP backbone that explicitly encodes the $SO(2)$ rotational symmetry of MAV dynamics. 
As shown in Fig.~\ref{fig:flowchart}(c), equivariant feature pairs (e.g., $(p_x,p_y)$, $(v_x,v_y)$, $(\omega_x,\omega_y)$, and selected attitude terms) rotate consistently under yaw, while invariant features (e.g., $p_z$, $v_z$, $\omega_z$) remain fixed. 
This structured decomposition ensures that the EMLP maintains rotation consistency where symmetry holds.

However, in multi-task learning, strict equivariance can impose excessive constraints: different maneuvers interact with symmetry in distinct ways, and heterogeneous reward shaping emphasizes task-specific features. 
Thus, while the equivariant actor provides a strong inductive bias, it must remain flexible enough to adapt to task asymmetries.

\subsection{Feature-wise Linear Modulation}

FiLM layers modulate state features using command inputs, parameterizing maneuvers through learned scaling and shifting operations:
\begin{equation}
    \mathrm{FiLM}(x, c) = x \gamma(c) + \beta(c),
\end{equation}
where $x$ denotes state features and $c$ denotes command features. 

In practice, FiLM modulation is applied before the features are organized into group representations. In GEAR, FiLM serves two purposes. First, it conditions the shared equivariant backbone on task commands, ensuring that a single actor can represent multiple maneuvers through lightweight modulation. Second, by introducing additional bias flexibility, FiLM allows the network to deviate from overly rigid equivariance when necessary, while still exploiting $SO(2)$ symmetry where it remains advantageous. This design provides a practical balance between symmetry-based inductive bias and task-specific adaptability.

\subsection{Multi-Head Critic for Multi-Task Value Estimation}

While FiLM addresses symmetry relaxation in the actor, the critic must faithfully capture reward structures that differ across the four aerobatic tasks. 
For example, Hover requires maintaining absolute position and orientation stability, Flip emphasizes accurate angular velocity tracking during vertical rotations, Rotate enforces consistent tangential motion around a target, and Roll focuses on precise tracking of commanded roll revolutions. 
These task-specific reward signals differ in both structure and sensitivity, making a single shared value head prone to conflating them and leading to inaccurate value estimation. We therefore adopt a multi-head critic design: a shared equivariant backbone extracts common geometric features, while separate value heads specialize in different tasks. 
This modularity prevents reward conflation, stabilizes multi-task optimization, and ensures that value estimation remains accurate across tasks with heterogeneous objectives.

\section{Experiments}\label{sec_experiments}

\subsection{Experimental Setup}

The MAV weighs 0.46\,kg, has a motor-to-motor span of 0.149\,m, and achieves a thrust-to-weight ratio of 4.1. 
A Vicon motion capture system provides position, attitude, and velocity measurements at 200\,Hz. 
The angular velocity controller operates at 1000\,Hz, while high-level policy inference is executed at 100\,Hz.

The SO(2)-equivariant actor is implemented using the open-source ESCNN library~\cite{cesa2022a}, which offers group representations and symmetry-aware architectural components. 
Policies are trained with Proximal Policy Optimization (PPO)~\cite{schulman_proximal_2017} on 2048 parallel environments in the IsaacGym platform~\cite{makoviychuk2021isaac}, requiring 8.9 hours on an RTX 4070 Ti GPU. 
During training, the actor receives noisy observations to enhance robustness, whereas the critic accesses ground-truth states. 
At the start of each episode, the initial states (including position, orientation, velocity, motor speeds, command inputs, and dynamic parameters) are randomized across environments. 
The randomization range expands progressively, forming a simple curriculum learning schedule. 
Episodes terminate when the MAV descends below a safety altitude or exceeds the maximum horizon.

\subsection{Experimental Results}

\subsubsection{Training Performance Comparison}
\begin{figure*}[t]
    \centering
    \includegraphics[width=0.99\textwidth]{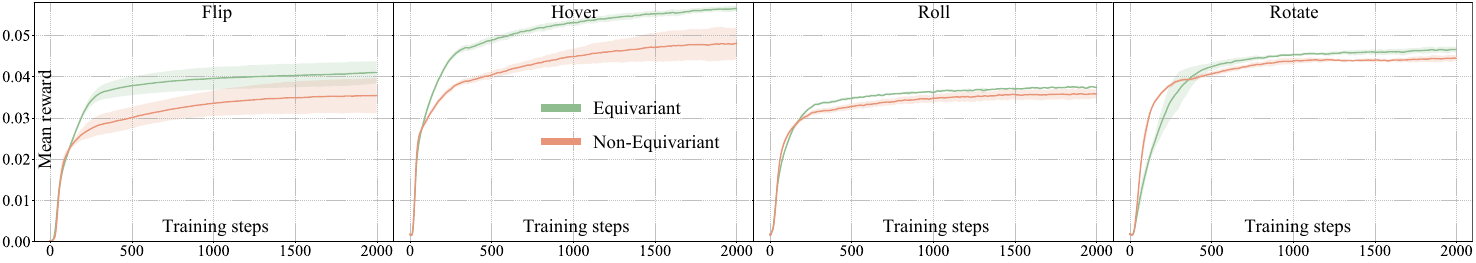}
\caption{Training curves of EMLP- and MLP-based frameworks on four tasks (Flip, Hover, Roll, and Rotate). Results are from a single joint training run of one policy across all tasks. Each curve represents the average reward over 10 random seeds, with shaded regions indicating the $95\%$ confidence interval. }
    \label{fig:Running_result}
\end{figure*}

We evaluate training performance using the individual task rewards. 

Fig.~\ref{fig:Running_result} shows the learning curves of EMLP and MLP frameworks in the multi-task setting, where one policy is jointly trained and evaluated on Flip, Hover, Roll, and Rotate. 

Across all tasks, the EMLP framework consistently outperforms the MLP baseline in both convergence speed and final performance. 
For Flip, EMLP converges faster and sustains higher rewards with smaller variance. 
In Hover and Roll, it achieves superior reward levels and more stable learning, as reflected by narrower confidence intervals. 
For Rotate, EMLP again shows robustness, maintaining a clear margin over MLP throughout training. 

These results demonstrate that encoding equivariance into the policy substantially improves multi-task reinforcement learning, leading to higher rewards, faster convergence, and reduced variance. This enhanced generalization and stability make the EMLP framework well suited for scalable and reliable learning of diverse aerobatic maneuvers.
\subsubsection{Ablation Study}
\begin{table*}[t]
\centering
\footnotesize
\setlength{\tabcolsep}{4pt}
\renewcommand{\arraystretch}{1.2}
\caption{Ablation study on network architecture and equivariance. 
Reported metrics include mean success rate (\%), SCD, and primary error terms, averaged over 4096 test episodes across four aerobatic tasks. 
\textbf{Baseline}: re-implementation of the SOTA MTRL framework for MAV~\cite{xing2024multi}. 
\textbf{Backbone Only}: plain backbone (MLP or EMLP), without FiLM or multi-head critic. 
\textbf{FiLM Only}: FiLM-based task conditioning, without multi-head critic. 
\textbf{Multi-Head Only}: multi-head critic with shared backbone, without FiLM. 
\textbf{Multi-Head+FiLM}: FiLM-based conditioning combined with a multi-head critic. 
\textbf{Multi-Head+IFiLM}: SO(2)-equivariant IFiLM module introduced in EquAct~\cite{zhu_equact_2025}. 
Best-performing results for each metric are highlighted in bold; for error metrics and SCD, lower values indicate better performance.}

\label{tab:ablation_performance}

\begin{tabularx}{\textwidth}{l l
    >{\centering\arraybackslash}X >{\centering\arraybackslash}X >{\centering\arraybackslash}X
    >{\centering\arraybackslash}X >{\centering\arraybackslash}X >{\centering\arraybackslash}X
    >{\centering\arraybackslash}X >{\centering\arraybackslash}X >{\centering\arraybackslash}X
    >{\centering\arraybackslash}X >{\centering\arraybackslash}X >{\centering\arraybackslash}X}
\toprule
\multirow{2}{*}{Network} & \multirow{2}{*}{Architecture}
& \multicolumn{3}{c}{Hover Task}
& \multicolumn{3}{c}{Flip Task}
& \multicolumn{3}{c}{Roll Task}
& \multicolumn{3}{c}{Rotate Task} \\
\cmidrule(lr){3-5} \cmidrule(lr){6-8} \cmidrule(lr){9-11} \cmidrule(lr){12-14}
& & Suc.R (\%) & SCD & Hover.Err (m)
  & Suc.R (\%) & SCD & Ang.Err (rad/s)
  & Suc.R (\%) & SCD & Roll.Err (rad)
  & Suc.R (\%) & SCD & Vel.Err (m/s) \\
\midrule

% MLP
\multirow{4}{*}{MLP} 
    & Backbone Only                & 98.34 & 0.0067 & 0.0733   & 58.11 & 6.0193 & 15.0911   & 1.37 & 6.3810 & 6.3821   & 98.14 & 0.0712 & 0.3716   \\
    & FiLM Only                & 98.54 & 0.0047 & 0.0748   & 34.08 & 9.4524 & 15.0001   & 1.56 & 6.3803 & 6.3819   & 99.90 & 0.0041 & 0.3597   \\
    & Multi-Head Only & 98.73 & 0.0048 & \textbf{0.0658}   & 89.36 & 0.7728 & 14.7990   & 66.44 & 1.9564 & 2.0257   & 100.00 & 0.0000 & 0.3055  \\
    & Multi-Head+FiLM   & 98.63 & 0.0038 & 0.0756   & 89.26 & 1.3385 & 17.0861   & 84.42 & 0.4790 & 0.5658   & 99.90 & 0.0068 & 0.5777   \\
        & Baseline & 99.71&0.0002&0.0830 &99.02 & 0.0764 & 13.1176 & 56.84&2.7592&2.8167 &81.05&0.0433&  \textbf{0.2246}    \\
\midrule
% EMLP
\multirow{5}{*}{EMLP}
    & Backbone Only                & 81.07 & 0.0427 & 0.1276   & 71.48 & 5.2821 & 28.0750   & 22.71 & 4.0200 & 4.0424   & 0.00 & 10.2812 & 10.2812   \\
    & FiLM Only               & 98.34 & 0.0017 & 0.1121   & 94.63 & 0.4072 & 10.3995   & 48.83 & 3.1630 & 3.2124   & 100.00 & 0.0000 & 0.3054   \\
    & Multi-Head Only & 52.63 & 0.0950 & 0.1432   & 48.54 & 11.8226 & 25.9429   & 45.76 & 3.3884 & 3.4348   & 58.98 & 1.1321 & 1.8313   \\
    & Multi-Head+IFiLM & 94.92 & 0.0059 & 0.1011   & \textbf{99.22} & \textbf{0.0561} & 10.2399   & 87.70 & 0.3363 & 0.4240   & 100.00 & 0.0000 & 0.2912   \\
    & \textbf{Multi-Head+FiLM} & \textbf{99.90} & \textbf{0.0001} & 0.0795 & 95.80 & 0.3900 & \textbf{9.4787} & \textbf{99.71} & \textbf{0.0085} & \textbf{0.1082} & \textbf{100.00} & \textbf{0.0000} & 0.2753 \\
\bottomrule
\end{tabularx}

\end{table*}
We evaluate model performance using three key metrics: success rate (SR), mean error, and Success-weighted by Command Distance (SCD). 
For success rate, we adopt task-specific criteria: Hover is successful if the final position error is less than $0.1$\,m and the orientation error is less than $10^\circ$; Flip is successful if the cumulative pitch change reaches $\pi$\,rad; Roll is successful if the final roll angle error after one commanded roll is less than $0.26$; Rotate is successful if the mean radius error around the desired $r=1.2$\,m trajectory is less than $0.1$\,m. 
SCD provides a comprehensive measure by capturing both maneuver success and command tracking. For each episode $i$, the Command Distance ($C_i$) is defined as the cumulative error between the intended command and the actual action executed by the MAV. The SCD is then computed as
\begin{equation}
    SCD = \frac{1}{N} \sum_{i=1}^{N} (1 - \mathbb{I}\{d_i \le \tau\}) C_i,
\end{equation}
where $\mathbb{I}\{d_i \le \tau\}$ is the indicator of successful completion, and $C_i$ represents the command tracking error. This design is particularly important for tasks such as Flip, where a policy may track angular velocity commands yet fail to complete the maneuver due to momentum loss.

All ablation experiments are conducted on the four representative aerobatic primitives (Hover, Flip, Roll, and Rotate) under randomized command conditions. 
For Hover, the target position and orientation are randomized with zero velocity. 
For Rotate, the commanded tangential velocity is uniformly sampled from $[-6,6]$\,m/s, with values $|v|>4$ unseen during training. 
For Roll, the commanded number of revolutions is sampled from $[-5,5]$, with $|n|>3$ representing unseen test commands. 
For Flip, the angular velocity command is sampled from $[2,8]$\,rad/s, while training is restricted to $[4,6]$\,rad/s. 
This setup enables evaluation of both in-distribution performance and generalization to out-of-distribution commands.

\textbf{Baseline comparison.} 
We re-implement the state-of-the-art MTRL framework for MAVs~\cite{xing2024multi} which employs a multi-critic architecture with shared task encoders under our aerobatic suite. 
It achieves an overall SR of 84.16\% but is heavily limited by poor Roll performance. 
Training also requires 10.1 hours, more than one hour longer than our framework. 
While competitive on Rotate velocity tracking and Flip, the baseline lags behind by 8.89\% compared with the MLP Multi-Head+FiLM variant 
and by 14.79\% compared with our GEAR. 
Due to the encoder design, which mixes observations before feeding them into both actor and critic, it is not possible to directly replace their backbone with an EMLP. 
This prevents a fair EMLP-based baseline comparison, and we therefore keep their original architecture for evaluation.

\textbf{Ablation results.} 
Table~\ref{tab:ablation_performance} summarizes the ablation results.The plain MLP Base achieves good performance in Hover but struggles in Flip and Roll, while the EMLP Base performs poorly overall, confirming that strict equivariance alone reduces policy expressiveness. \textbf{FiLM only} improves performance by conditioning on command inputs and helps on simpler tasks, but it fails to learn the more difficult Roll maneuver, indicating that modulation without value disentanglement is insufficient for highly nonlinear dynamics. In contrast, Multi-Head only stabilizes optimization by separating reward signals, leading to more balanced performance across tasks. However, due to the absence of task conditioning, its expressiveness remains limited. Replacing FiLM with \textbf{Multi-Head+IFiLM}, introduced in EquAct~\cite{zhu_equact_2025}, enforces stronger invariance but sacrifices robustness in roll. The best overall performance comes from \textbf{Multi-Head+FiLM}, which achieves near-perfect success rates and the lowest errors, highlighting the benefit of combining equivariance with adaptive conditioning and modular value estimation. These gains are especially pronounced in EMLP-based models: because stronger equivariance constraints reduce flexibility, FiLM and Multi-Head become essential to relax overly rigid symmetry and disentangle heterogeneous rewards. This evidence directly validates our conjecture in the method section. Furthermore, our framework maintains high success rates on unseen commands, demonstrating strong generalization ability.

\subsection{Real World Experiment}
\begin{figure*}[t]
    \centering
    \includegraphics[width=1\textwidth]{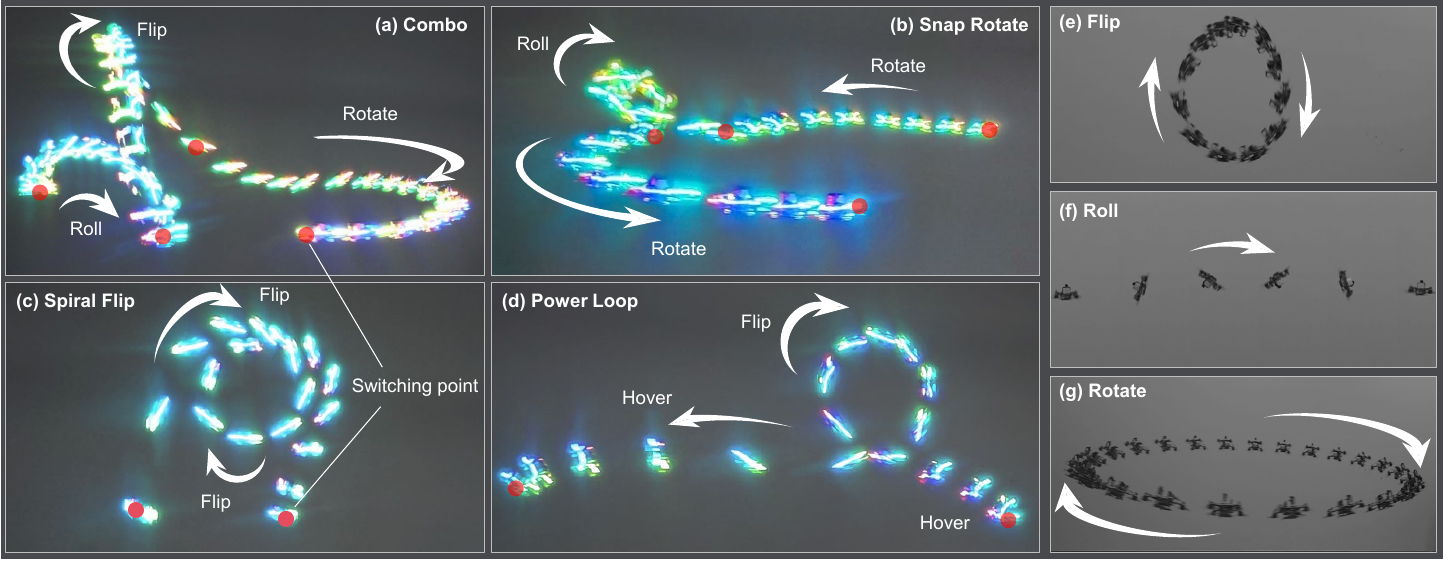}
    \caption{Demonstration of real-world acrobatic maneuvers. Arrows denote the direction of motion. For the Roll maneuver, five consecutive images are concatenated to highlight its nearly constant altitude. Red markers indicate the timing of issued commands.}
    \label{fig:flowchart}
\end{figure*}

In this section, we validate the transfer of our learned primitives to real-world platforms and demonstrate that complex aerobatic maneuvers can be composed by manually selecting task types and motion parameters. Commands are issued through three mechanisms: (i) sequential triggering once the previous maneuver is completed, (ii) timed triggering after a predefined interval, and (iii) manual triggering through a human–machine interface.

For the \textit{Combo} maneuver, a Roll command is followed after $0.1$~s by a Flip at $5~\text{rad/s}$, then a Rotate command after $0.6$~s with the specified center, and finally a Hover command after $1$~s. For the \textit{Snap Rotate}, the MAV begins with a Rotate command at a tangential velocity of $3~\text{m/s}$ around a specified center point. After $0.5$~s, a single-Roll command is issued. Once the Roll is completed, the same Rotate command is resumed, and the MAV continues the rotation until it reaches the hover waypoint. For the \textit{Spiral Flip}, the MAV is commanded to perform two consecutive Flips at $5~\text{rad/s}$ around a specified point, after which a Hover command is applied to stabilize the vehicle. To realize the \textit{Power Loop}, the MAV is initialized with a Flip command at an angular velocity of $5~\text{rad/s}$. Then the maneuver is terminated with a Hover command at the designated position. 

These results confirm that the proposed framework enables reliable composition of learned primitives into diverse and complex aerobatic maneuvers on real hardware with simple waypoints.

\section{Conclusion}

We introduced GEAR, a framework for multi-task reinforcement learning in drone aerobatics that integrates SO(2)-equivariant architectures. By combining geometric inductive bias with adaptive conditioning, GEAR enables a single policy to perform diverse high-speed maneuvers with greater efficiency, stability, and generalization, achieving clear improvements over conventional baselines. This approach advances autonomous aerobatic flight and offers a blueprint for broader multi-task robotic control.

\bibliographystyle{IEEEtran}
\bibliography{IEEEabrv, capture_references}

\begin{thebibliography}{10}
\providecommand{\url}[1]{#1}
\csname url@rmstyle\endcsname
\providecommand{\newblock}{\relax}
\providecommand{\bibinfo}[2]{#2}
\providecommand\BIBentrySTDinterwordspacing{\spaceskip=0pt\relax}
\providecommand\BIBentryALTinterwordstretchfactor{4}
\providecommand\BIBentryALTinterwordspacing{\spaceskip=\fontdimen2\font plus
\BIBentryALTinterwordstretchfactor\fontdimen3\font minus \fontdimen4\font\relax}
\providecommand\BIBforeignlanguage[2]{{%
\expandafter\ifx\csname l@#1\endcsname\relax
\typeout{** WARNING: IEEEtran.bst: No hyphenation pattern has been}%
\typeout{** loaded for the language `#1'. Using the pattern for}%
\typeout{** the default language instead.}%
\else
\language=\csname l@#1\endcsname
\fi
#2}}
\renewcommand\BIBentryALTinterwordstretchfactor{4}

\bibitem{kaufmann2020RSS}
E.~Kaufmann, A.~Loquercio, R.~Ranftl, M.~M{\"u}ller, V.~Koltun, and D.~Scaramuzza, ``Deep drone acrobatics,'' in \emph{Proceedings of Robotics: Science and Systems}, Corvalis, Oregon, USA, July 2020.

\bibitem{wang_unlocking_2025}
\BIBentryALTinterwordspacing
M.~Wang, \emph{et~al.}, ``Unlocking aerobatic potential of quadcopters: Autonomous freestyle flight generation and execution,'' \emph{Science Robotics}, vol.~10, no. 101, p. eadp9905, 2025. [Online]. Available: \url{https://www.science.org/doi/abs/10.1126/scirobotics.adp9905}
\BIBentrySTDinterwordspacing

\bibitem{han2025reactive}
Z.~Han, \emph{et~al.}, ``Reactive aerobatic flight via reinforcement learning,'' \emph{IEEE Robotics and Automation Letters}, vol.~10, no.~10, pp. 11\,014--11\,021, 2025.

\bibitem{kaufmann_champion-level_2023}
E.~Kaufmann, L.~Bauersfeld, A.~Loquercio, M.~Müller, V.~Koltun, and D.~Scaramuzza, ``Champion-level drone racing using deep reinforcement learning,'' \emph{Nature}, vol. 620, no. 7976, pp. 982--987, 2023.

\bibitem{song2023reaching}
Y.~Song, A.~Romero, M.~M{\"u}ller, V.~Koltun, and D.~Scaramuzza, ``Reaching the limit in autonomous racing: Optimal control versus reinforcement learning,'' \emph{Science Robotics}, vol.~8, no.~82, p. eadg1462, 2023.

\bibitem{wang_aerobatic_2022}
Z.~Wang, R.~Gros, and S.~Zhao, ``Aerobatic {Tic}-{Toc} {Control} of {Planar} {Quadcopters} via {Reinforcement} {Learning},'' \emph{IEEE Robotics and Automation Letters}, vol.~7, no.~2, pp. 2140--2147, Apr. 2022.

\bibitem{yin_taco_2025}
Z.~Yin, C.~Zheng, S.~Guo, Z.~Wang, and S.~Zhao, ``{TACO}: {General} {Acrobatic} {Flight} {Control} via {Target}-and-{Command}-{Oriented} {Reinforcement} {Learning},'' Mar. 2025, arXiv:2503.01125 [cs].

\bibitem{xing2024multi}
J.~Xing, I.~Geles, Y.~Song, E.~Aljalbout, and D.~Scaramuzza, ``Multi-task reinforcement learning for quadrotors,'' \emph{IEEE Robotics and Automation Letters}, vol.~10, no.~33, p. 2112–2119, Mar 2025.

\bibitem{bauersfeld_user-conditioned_2023}
L.~Bauersfeld, E.~Kaufmann, and D.~Scaramuzza, ``User-conditioned neural control policies for mobile robotics,'' in \emph{2023 IEEE International Conference on Robotics and Automation (ICRA)}, 2023, pp. 1342--1348.

\bibitem{zhu_equact_2025}
X.~Zhu, Y.~Qi, Y.~Zhu, R.~Walters, and R.~Platt, ``Equact: An se (3)-equivariant multi-task transformer for open-loop robotic manipulation,'' \emph{arXiv preprint arXiv:2505.21351}, 2025.

\bibitem{cesa2022a}
G.~Cesa, L.~Lang, and M.~Weiler, ``A program to build {E(N)}-equivariant steerable {CNN}s,'' in \emph{International Conference on Learning Representations}, 2022.

\bibitem{huang2023leveraging}
H.~Huang, D.~Wang, A.~Tangri, R.~Walters, and R.~Platt, ``Leveraging symmetries in pick and place,'' \emph{The International Journal of Robotics Research}, vol.~43, no.~4, pp. 550--571, 2023.

\bibitem{yu_equivariant_2025}
B.~Yu and T.~Lee, ``Equivariant reinforcement learning frameworks for quadrotor low-level control,'' \emph{IEEE Transactions on Control Systems Technology}, pp. 1--14, 2025.

\bibitem{pertigkiozoglou2024improving}
S.~Pertigkiozoglou, E.~Chatzipantazis, S.~Trivedi, and K.~Daniilidis, ``Improving equivariant model training via constraint relaxation,'' \emph{Advances in Neural Information Processing Systems}, vol.~37, pp. 83\,497--83\,520, 2024.

\bibitem{NEURIPS2023_c95c0496}
J.~Brehmer, J.~Bose, P.~de~Haan, and T.~S. Cohen, ``Edgi: Equivariant diffusion for planning with embodied agents,'' in \emph{Advances in Neural Information Processing Systems}, vol.~36, 2023, pp. 63\,818--63\,834.

\bibitem{wang_practical_2025}
D.~Wang, B.~Hu, S.~Song, R.~Walters, and R.~Platt, ``A practical guide for incorporating symmetry in diffusion policy,'' \emph{arXiv preprint arXiv:2505.13431}, 2025.

\bibitem{sun_comparative_2022}
S.~Sun, A.~Romero, P.~Foehn, E.~Kaufmann, and D.~Scaramuzza, ``A {Comparative} {Study} of {Nonlinear} {MPC} and {Differential}-{Flatness}-{Based} {Control} for {Quadrotor} {Agile} {Flight},'' \emph{IEEE Transactions on Robotics}, vol.~38, no.~6, pp. 3357--3373, Dec. 2022.

\bibitem{zhao2025RLBook}
S.~Zhao, \emph{Mathematical Foundations of Reinforcement Learning}.\hskip 1em plus 0.5em minus 0.4em\relax Springer Nature Press, 2025.

\bibitem{UNMANNED}
\BIBentryALTinterwordspacing
J.~Fu and F.~Yang, ``Application of reinforcement learning in uav tasks: A survey,'' \emph{Unmanned Systems}, vol.~0, no.~0, pp. 1--14, 0. [Online]. Available: \url{https://doi.org/10.1142/S2301385026300015}
\BIBentrySTDinterwordspacing

\bibitem{simeonov2022neural}
A.~Simeonov, \emph{et~al.}, ``Neural descriptor fields: {SE}(3)-equivariant object representations for manipulation,'' in \emph{2022 International Conference on Robotics and Automation (ICRA)}.\hskip 1em plus 0.5em minus 0.4em\relax IEEE, 2022, pp. 6394--6400.

\bibitem{wang2022so}
\BIBentryALTinterwordspacing
D.~Wang, R.~Walters, and R.~Platt, ``{SO}(2)-equivariant reinforcement learning,'' in \emph{International Conference on Learning Representations}, 2022. [Online]. Available: \url{https://openreview.net/forum?id=ax8O8n1g5G}
\BIBentrySTDinterwordspacing

\bibitem{su_leveraging_2024}
Z.~Su, \emph{et~al.}, ``Leveraging symmetry in rl-based legged locomotion control,'' in \emph{2024 IEEE/RSJ International Conference on Intelligent Robots and Systems (IROS)}, 2024.

\bibitem{ordonez-apraez_morphological_2025}
D.~O. Apraez, \emph{et~al.}, ``Morphological symmetries in robotics,'' \emph{The International Journal of Robotics Research}, vol.~0, no.~0, p. 02783649241282422, 0.

\bibitem{smith_so2-equivariant_2024}
H.~Smith, A.~Shankar, J.~Gielis, J.~Blumenkamp, and A.~Prorok, ``{SO}(2)-{Equivariant} {Downwash} {Models} for {Close} {Proximity} {Flight},'' \emph{IEEE Robot. Autom. Lett.}, vol.~9, pp. 1174--1181, Feb. 2024.

\bibitem{yu_equivariant_2023}
B.~Yu and T.~Lee, ``Equivariant reinforcement learning for quadrotor uav,'' in \emph{2023 American Control Conference (ACC)}, 2023, pp. 2842--2847.

\bibitem{10054413}
J.~Huang, \emph{et~al.}, ``Symmetry-informed reinforcement learning and its application to low-level attitude control of quadrotors,'' \emph{IEEE Transactions on Artificial Intelligence}, vol.~5, no.~3, pp. 1147--1161, 2024.

\bibitem{schulman_proximal_2017}
J.~Schulman, F.~Wolski, P.~Dhariwal, A.~Radford, and O.~Klimov, ``Proximal policy optimization algorithms,'' \emph{arXiv preprint arXiv:1707.06347}, 2017.

\bibitem{makoviychuk2021isaac}
V.~Makoviychuk, \emph{et~al.}, ``Isaac gym: High performance gpu-based physics simulation for robot learning,'' 2021.

\end{thebibliography}

\vfill

\end{document}